\documentclass[pdflatex,sn-mathphys]{sn-jnl}

\theoremstyle{thmstyleone}%
\theoremstyle{thmstyletwo}%

\theoremstyle{thmstylethree}%

\usepackage{setspace} 
\usepackage[pagewise,modulo]{lineno} 

\begin{document}

\title[\small Reliability of Robotic Ultrasound Scanning for Scoliosis Assessment]{\bf Reliability of Robotic Ultrasound Scanning for Scoliosis Assessment in Comparison with Manual Scanning}

\author*{\fnm{Maria} \sur{Victorova}}

\author{\fnm{Heidi Hin Ting} \sur{Lau}}

\author{\fnm{Timothy Tin-Yan} \sur{Lee}}

\author{\fnm{David} \sur{Navarro-Alarcon}}

\author{\fnm{Yongping} \sur{Zheng}}

\affil{\orgname{The Hong Kong Polytechnic University}, \orgaddress{ \country{Hong Kong}}}


\abstract{\normalsize {{\bf{Background:}} Ultrasound (US) imaging for scoliosis assessment is challenging for a non-experienced operator. The robotic scanning was developed to follow a spinal curvature with deep learning and apply consistent forces to the patient's back.

{\bf{Methods:}}
23 scoliosis patients were scanned with US device both, robotically and manually. Two human raters measured each subject's spinous process angles (SPA) on robotic and manual coronal images. 

{\bf{Results:}}
The robotic method showed high intra- $(ICC >0.85)$ and inter-rater $(ICC >0.77)$ reliabilities. Compared with the manual method, the robotic approach
showed no significant difference ($p<0.05$) when measuring coronal deformity angles. The MAD for intra-rater analysis lies within an acceptable range from $0\degree$ to $5\degree$ for the minimum of $86\%$ and maximum $97\%$ of a total number of the measured angles.

{\bf{Conclusions:}}
This study demonstrated that scoliosis deformity angles measured on ultrasound images obtained with robotic scanning are comparable to those obtained by manual scanning. 
}

}


\maketitle

\newpage
\section{Introduction}

Quantitative assessment of curve severity for adolescent idiopathic scoliosis (AIS) patients is vital for progression monitoring and treatment planning.
The gold standard of scoliosis curve assessment is Cobb's angle \cite{cobb1948outline} on coronal X-ray. 
Although it has been proved that X-ray radiation causes higher risks of breast cancer \cite{hoffman1989breast} regular check-up of scoliosis is suggested for at least six months intervals \cite{levy1994projecting}. 
To reduce the radiation exposure caused by X-ray assessment, ultrasound methods for scoliosis assessment have been developed
\cite{Zheng2016,wang2015reliability,brink2018reliability,chen2012ultrasound}.

\begin{figure}[]
    \centering
    \includegraphics[width=\linewidth]{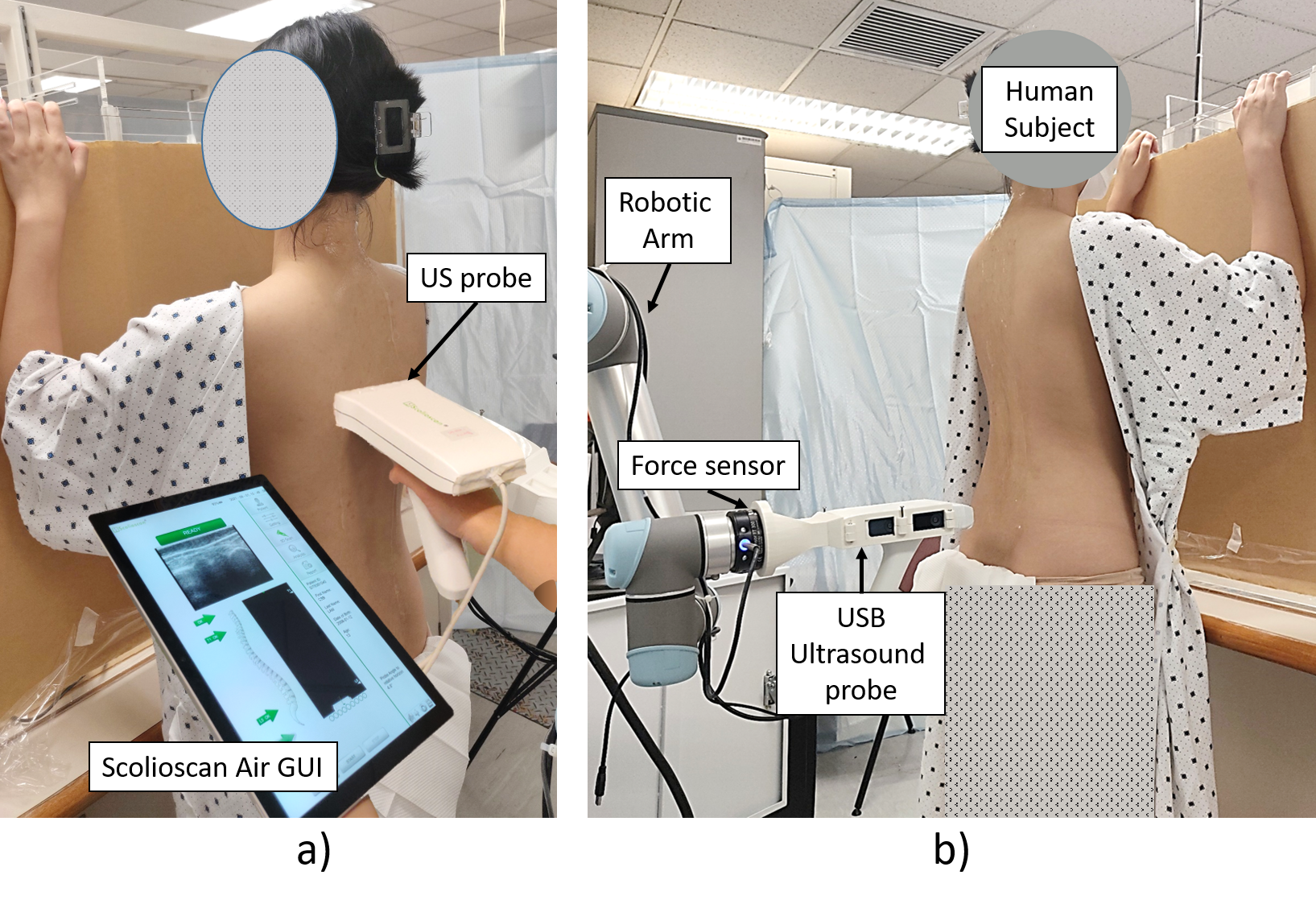}
    \caption[Setup for robotic scoliosis assessment for subject's robotic scanning.]{Setup for ultrasound scoliosis assessment: a) manual approach using Scolioscan Air system comprising of tablet PC with reconstruction software and USB ultrasound probe b) robotic approach using robotic arm UR5, force sensor FT300 and USB ultrasound probe similar to the Scolioscan Air system.}
    \label{fig:setup_GH}
\end{figure}

A portable 3D ultrasound system Scolioscan Air \cite{scolioAirlai2021} was designed for scoliosis curve assessment and progression monitoring for patients in upright position, resting arms on supporting wall with the elbows at 90 degrees. The system presented in Figure \ref{fig:setup_GH}a uses a palm-sized ultrasound probe with an optical 3D tracking device and a tablet PC with reconstruction software.
Using ultrasound B-mode spinal images with respective spatial information the Scolioscan software reconstructs the 3D spinal curve and generates coronal images for scoliosis angle measurements using the volume projection image (VPI) method \cite{Cheung2015}. With a non-planar re-slicing approach, where the skin surface is the reference for a slicing plane, the VPI method obtains an averaged intensity of all voxels of the volumetric image at a specified depth to form an image in the coronal plane. Jiang et al., \cite{Jiang2019} developed a 9-times faster coronal image generation approach than original VPI method by skipping the unnecessary 3D reconstruction and directly mapping the B-mode images and their corresponding positional data. Another method suggested by Vo et al., \cite{vo2019semi} creates the 3D spine reconstruction compounded from segmented bony features, which allows them to choose the plane of maximum curvature for coronal image projection.

There are several methods to measure scoliosis angle on ultrasound coronal images, depending on which anatomical feature is used for the measurement, such as center of lamina \cite{wang2015reliability}, spinous process \cite{Zheng2016} or transverse process \cite{brink2018reliability}. For Scolioscan system deformity angle measurements were based on spinous process. Zheng et al.,\cite{Zheng2016} showed that the spinous process angle (SPA) measured on US images had moderate to strong correlation with Cobb's angle measured on X-ray images. SPA had high intra-rater and inter-rater reliabilities with ICC higher than 0.9 and 0.87, respectively.
Another study supports this observation \cite{brink2018reliability} for Scolioscan system, excellent linear correlation ($R^2 = 0.97$) was observed between the SPA and Cobb's angle with high intra-rater and inter-rater reliability ICC was 0.97 and 0.95, respectively.
The validation study for Scolioscan Air showed the angles measured on coronal images obtained with Scolioscan \cite{Zheng2016} and portable version Scolioscan Air had a strong correlation, with $R^2$ higher than 0.7, and with intra-rater and inter-rater reliabilities with ICC larger than 0.94 and 0.88 respectively.

Ultrasonography is a safe and reliable technology for scoliosis evaluation \cite{Zheng2016,scolioAirlai2021,brink2018reliability}; however, due to the nature of ultrasound imaging, the inherent speckle (signal-dependent) noise affects the image quality, making it challenging for operators to distinguish anatomical characteristics.
The operator needs to adjust the time gain compensation (TGC) setting, then move the probe upwards, follow the spinal curve, apply sufficient pressure and change the probe's orientation normal to the human's back, at the same time observing the captured B-mode image quality, which is a non-trivial task for a non-experienced user. Usually, it takes around 3-6 months for an operator to become experienced. The poor scanning behavior, inconsistent contact with skin, and jerky movements would directly reduce the quality of the coronal images. This problem can be solved by using the robotic arm with a force sensor to apply a constant force to the human back and real-time adjust the probe's orientation to keep its surface normal to the skin. 

To reduce the effect of the human factor on spinal ultrasound examinations, several research groups proposed using robotic ultrasound systems \cite{Victorova2019,Tirindelli2020,Huang2021,li2021image}.
While several robotic-ultrasound systems are available for spinal applications and a few for scoliosis are developed, they primarily focus on proof of concept, showing the results obtained from a phantom or only involving a small sample size of human subjects. No extensive study has been conducted to compare the performance of robotic ultrasound assessment for spinal assessment with standard manual ultrasound procedures. 
Li et al., \cite{li2021image} presented a robotic approach for ultrasound imaging-based navigation to locate spinal features using deep learning (DL) and reinforcement learning on dataset of 648 images of spinous process to imitate the human decision-making during US scanning. The design was evaluated in an ultrasound simulation environment, and the resulted average error on the test set was $9.5\pm10.0mm$.
Another study \cite{Tirindelli2020} proposed a robotic-ultrasound system for spinal levels counting and injection site localization based on US images DL processing. This study uses 3972 images for training, resulting in an average localization accuracy of $2.1\pm2.6mm$ on the test set. Although these papers do not address scoliosis, the approaches employed can be used for various spinal applications. The following two studies directly address the scoliosis application. 
Yang et al. \cite{Huang2021} proposed the usage of the robotic arm to navigate through the spine based on the pre-calculated trajectory yield from the tracking camera image DL processing, using 2500 images for training. The scanning was done in a prone position, different from the Scolioscan Air protocol. The validation was performed on the phantom with different curvatures; the mean absolute error was 1.6 for moderate scoliosis. The concept was demonstrated on two human subjects, but the more extensive human study was not yet reported. The human scanning was performed for 120 seconds capturing 150 ultrasound frames per each.

The robotic ultrasound system used in this study (Figure \ref{fig:setup_GH}b) was designed for scoliosis assessment and was initially presented in \cite{Victorova2019} and extended in \cite{victorova2021follow}. It uses a DL-based navigation control to detect the center of the spine, which is defined as spinous process tip, in real-time at each B-mode ultrasound frame; thus, there is no need for additional external sensors, such as a camera. The proposed robotic system can also adjust the probe orientation to keep the ultrasound probe normal to the surface and maintain good acoustic coupling with the back of the subject. Good acoustic coupling can be defined as consistent contact with the human skin, where there is no air gap between the probe and skin, which would affect the resulting image quality. The details for the proposed robotic ultrasound system are described in section \ref{sec:robotic_scanning}. The dataset consisted of 25,774 images of spinous processes, and the resulting mean localization error for the human subjects test set was in the range $0.8-3.3mm$.

\begin{figure}[]
    \centering
    \includegraphics[width=\linewidth]{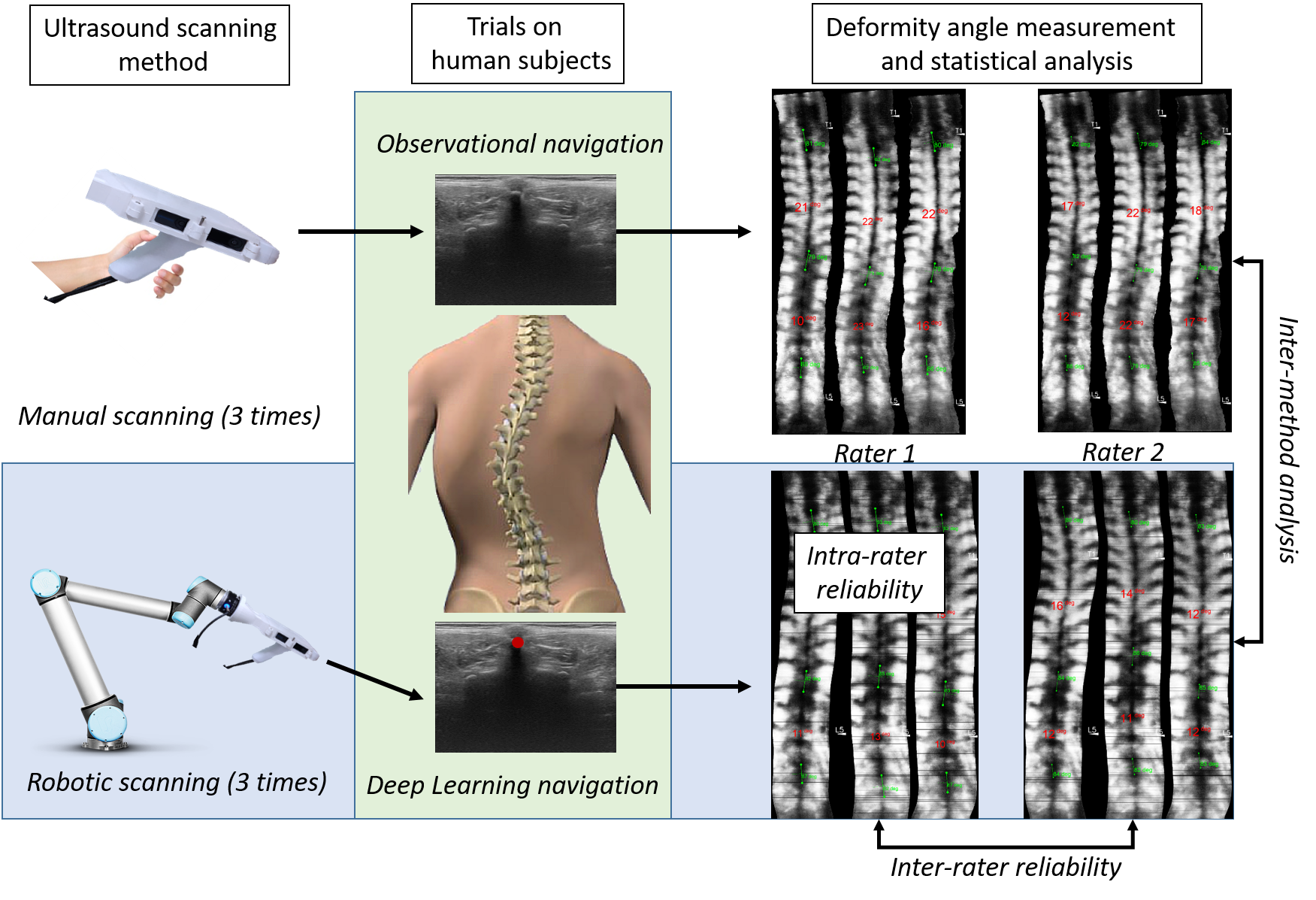}
    \caption[]{The study workflow diagram. Ultrasound scanning was performed with two methods: robotic and manual, three times each. During the robotic procedure the spinal navigation were performed with Deep Learning
     (DL) approach, whereas during manual procedure the navigation relies on human expertise in reading US images. Resulted coronal images of reconstructed spine were measured to assess the scoliosis severity by two human raters. The various analysis on reliability and validity were performed.}
    \label{fig:flow_diagram}
\end{figure}

The contribution of this study is to demonstrate that robotic scanning for 3D ultrasound imaging can achieve the same assessment for scoliosis in comparison with manual scanning.
 Figure \ref{fig:flow_diagram} shows the study flow chart. 
 We first collected human subject profiles and explored whether robotic scanning is beneficial and can replace manual scanning.
Then, statistical analysis was conducted 
for inter- and intra-rater reliability of scoliosis angle measurement, with ICC based on a single-rating
(k = 3), absolute-agreement, 2-way mixed-effects model. The mean absolute difference (MAD) of angles measured on images obtained by different methods would be compared with those reported previously.
 Two human raters measured spinal process angles on coronal images of six scans for each subject, three robotic and three manual scans.
 

To the best of our knowledge, this is the first study focusing on comparing robotic ultrasound scanning with human manual scanning for scoliosis assessment on a group of human subjects. In addition, it is the first work using a robotic ultrasound system which includes evaluation of robotic approach based on patient's comfortability survey.

\section{Methods}

\subsection{Subjects}
Twenty-three adolescents (21 female, 2 male) with confirmed or suspected scoliosis were recruited for this study. The main inclusion criteria were: mild or moderate scoliosis, age 14-18 years; no previous interventions on spine, since it may result in unpredictable image artifacts; no overweight ($BMI < 25.0 kg/m^2$) since a high BMI may result in poor ultrasound image quality with the 7.5 MHz probe utilized in this investigation.
Ethical approval was granted for this study\footnote{Ethical approval  HSEARS20210417002  was given by  Departmental Research Committee (on behalf of PolyU Institutional Review Board)}.

The subject leaned on the wall or wall-imitating stand during the scanning procedure. The subjects lifted their hands so that the angles between the elbow and the wall, elbow, and body are 90 degrees, as in Figure \ref{fig:setup_GH}. The subject was asked not to move but to breathe freely during the procedure.
Total six scans were performed on each subject, three times with manual approach and three times with robotic scanning. 
Before each scan, the subject was asked to relax and then stand in scanning position so that each scan would be independent of the previous; thus, the posture might vary slightly.
Of all the subjects, ten chose to participate in the survey on the comfortability of robotic and manual scanning.

\subsection{Manual Scanning Procedure}

The manual scoliosis assessment was performed with a Scolioscan Air system, consisting of an ultrasound transducer of 80 mm aperture and frequency of 7.5 MHz (UW-1C, Sonoptech, Beijing, China), a Realsense depth camera (T265, Intel, USA), and a tablet PC. The software installed on the PC collects the ultrasound and spatial data, processes it to form 3D reconstruction and coronal spinal images, and measures the deformity angle. The spatial data of the ultrasound device is obtained through a camera attached to the probe using a vision/odometry-based vSLAM algorithm \cite{Karlsson2005_slam}. This algorithm tracks the changes in the environment (due to the probe motion during the scanning) from the initial position and outputs the position and orientation of the tracking camera.

 For the assessment procedure, the operator of the ultrasound probe set the initial (at the level below L5) and final (around C5 of the cervical spine) points of the scanning and then started to move the probe upwards, applying enough pressure to maintain the stable probe-skin contact to see the bony features. The operator tried their best to keep the spinous process in the middle of the ultrasound frames available in real-time for observation. The average time to complete a scan was around 60 seconds. The scanning speed could be further increased, but since the robotic system (due to the hardware limitations) receives images and operates at the frequencies of 30 fps (as opposed to the Scolioscan Air system - 60 fps), the speed of the robotic scan has to be twice slower than the manual. Since the usual scanning speed is around 8mm/s-10mm/s, the robotic scan was performed with 4mm/s.

The manual ultrasound scanning of the human back was performed three times by an experienced operator with Scolioscan Air \cite{scolioAirlai2021}. After each scan, nine coronal images at different slicing depths were generated for the operator to visually assess the quality of the scan and measure the spinal curvature to generate the on-site report for the subject's use.

\subsection{Robotic Scanning Procedure}
\label{sec:robotic_scanning}

The setup for robotic scoliosis assessment, shown in Figure \ref{fig:setup_GH}b, employs a robotic arm (Universal robot UR5) with a force sensor affixed (FT300, Robotiq) and a palm-sized ultrasound probe (Sonoptek, Beijing).
The linear ultrasound probe with a width of 80 mm has a frequency of 7.5 MHz with an imaging depth of 6 cm and transfers raw data at ten frames per second via a USB port to a PC, where the images are produced in 640x480 pixels.
The robot was connected to a PC through TCP/IP protocol, and the control algorithm was launched at a 30 fps Hz rate.

\begin{figure}[]
    \centering
    \includegraphics[width=\linewidth]{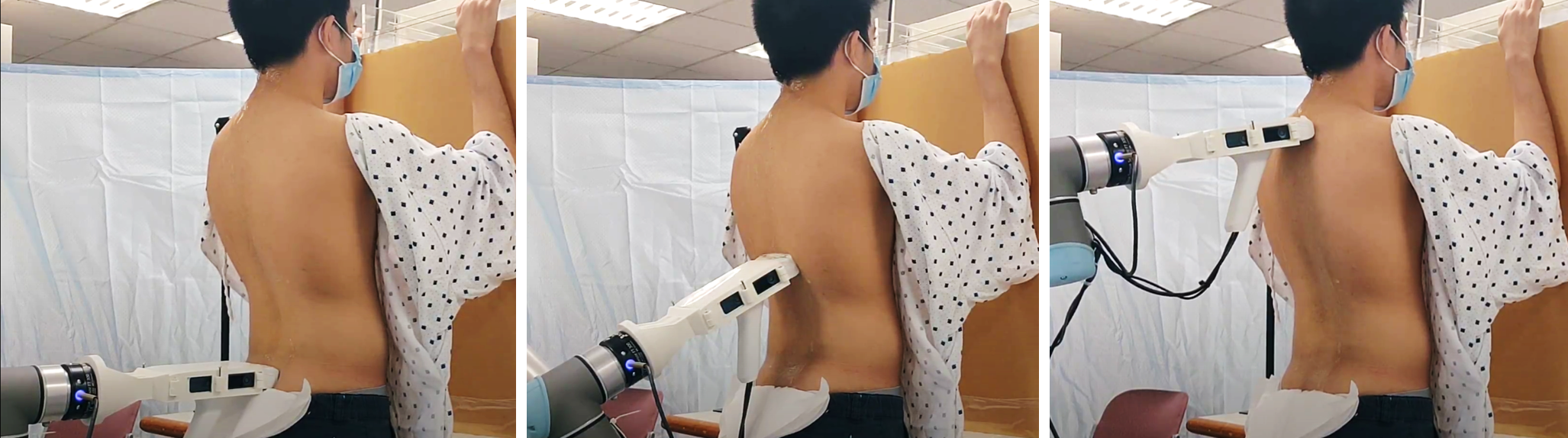}
    \caption[]{The images from the robotic scanning procedure video for one of the subjects. The robot adjusts the ultrasound probe rotation according the force sensing and smoothly scans through the patient's back, maintaining constant contact.}
    \label{fig:robotic_scanning_snapshots}
\end{figure}

The patient's posture is similar to the Scolioscan Air scanning procedure. The operator drags the robotic actuated probe using build-in admittance control to the base of the subject's spine, with the probe parallel to the ground. From this position, the robot starts to drive the ultrasound probe further until it reaches the human skin (force detected). Then the robot starts moving upwards, applying the preset force (which was chosen in the range of 10-15N, according to the subject's BMI, as discussed in \cite{Tirindelli2020}), adjusting rotations, and following the curve by detecting spinous processes in real-time. Figure \ref{fig:robotic_scanning_snapshots} illustrates some of the timestamps of the robotic procedure, showing how the robotic arm follows the profile of the human back, adjusting the probe's orientation. The average scanning time for one scan was around 120 seconds.

\begin{figure}[]
    \centering
    \includegraphics[width=\linewidth]{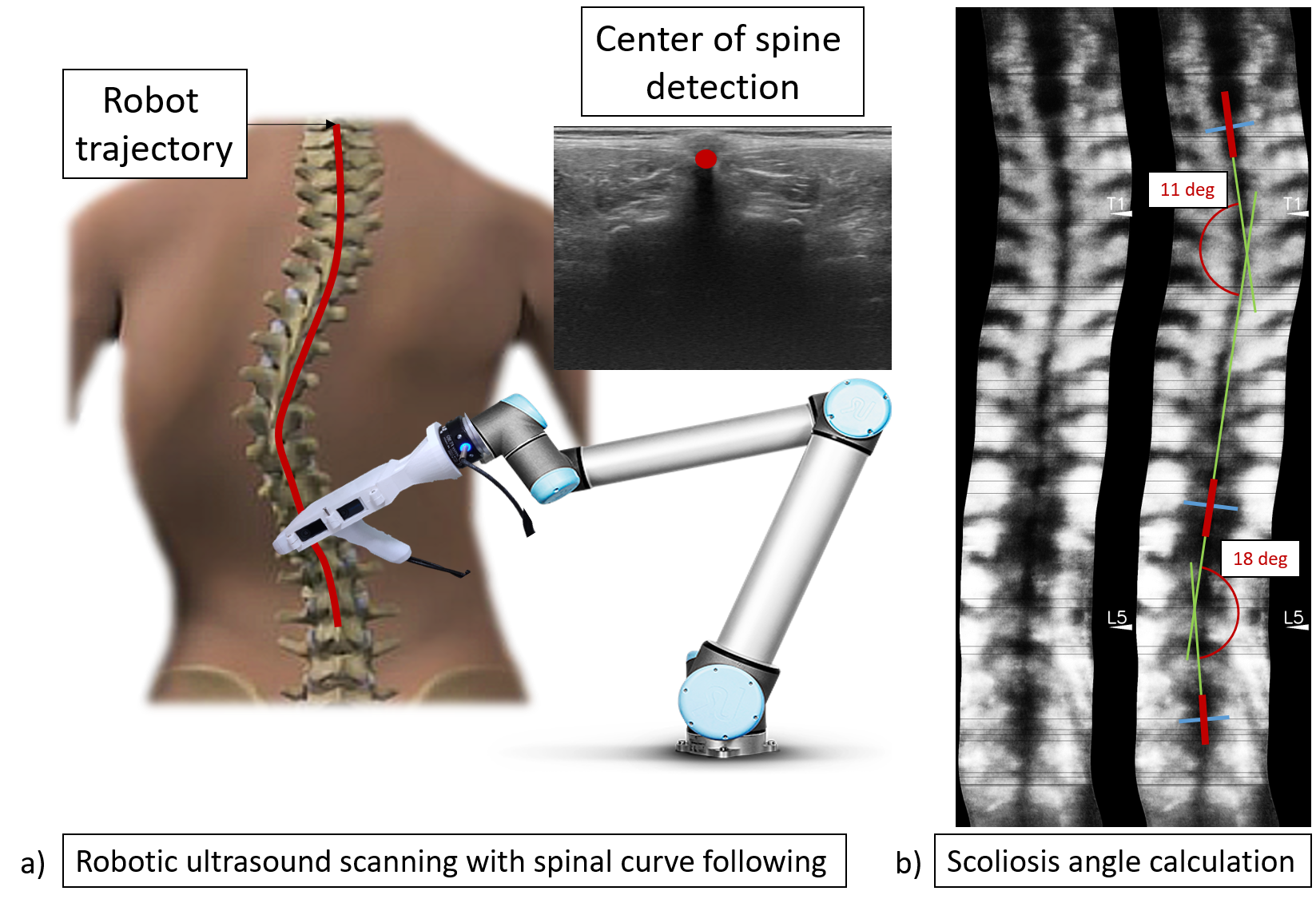}
    \caption[Setup for robotic scoliosis assessment for subject's robotic scanning.]{a) Overview of the robotic scanning approach, where the robotic arm automatically follows the spinal curvature based on B-mode images; b) Principles of spinous process angle (SPA) measurement based on the indication of the most tilted vertebrae on the coronal image of spinal reconstruction.}
    \label{fig:robotic_appr_angle}
\end{figure}

\begin{figure}[]
    \centering
    \includegraphics[width=\linewidth]{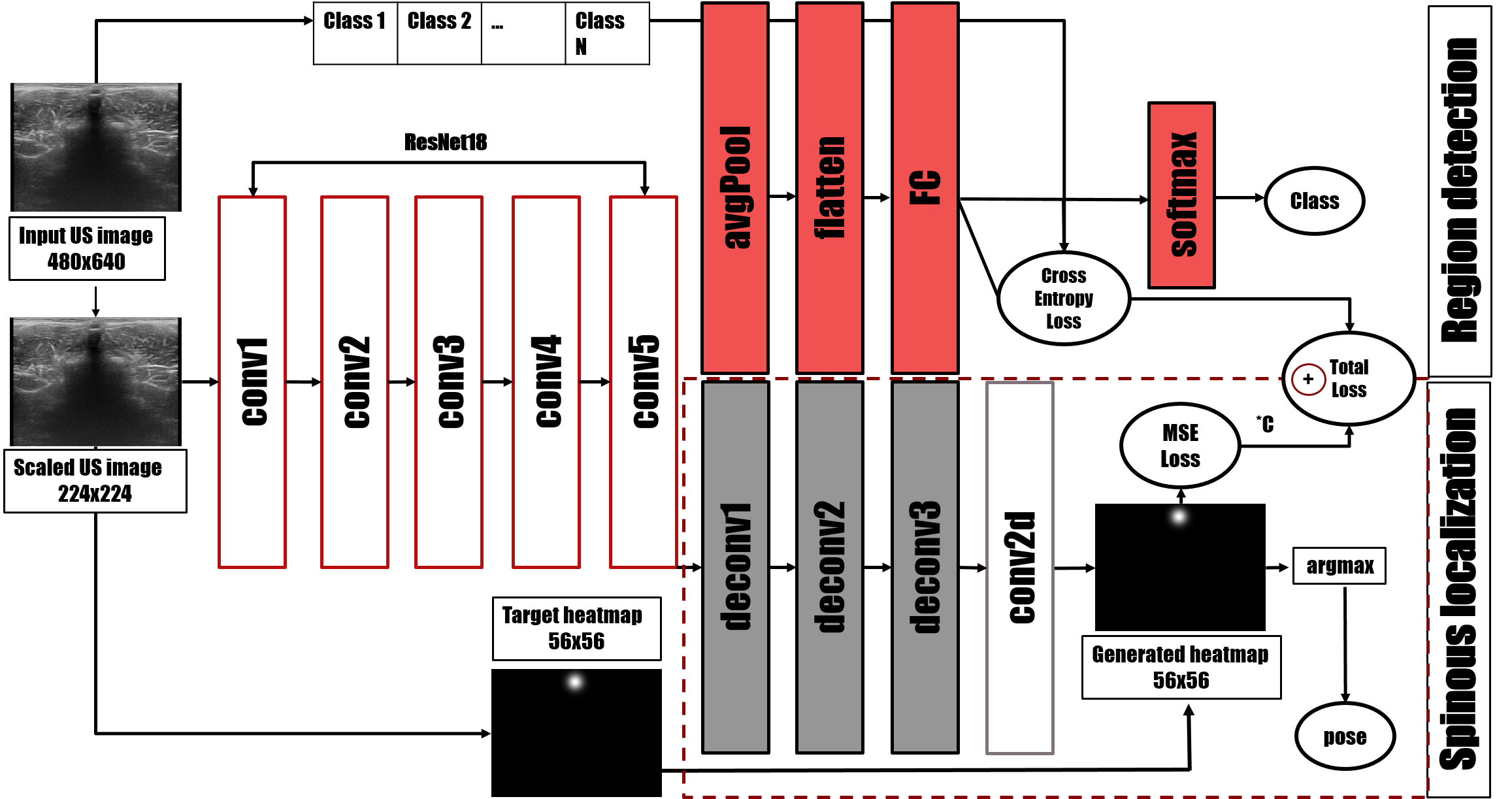}
    \caption[]{Network for spinous process localization and spinal region classification.}
    \label{fig:network_two_heads}
\end{figure}

\textit{Robotic control method.} The schematic overview of the robotic scanning approach is presented in Figure \ref{fig:robotic_appr_angle}a. 
The robotic arm is controlled to maintain the constant force applied to the subjects back with PID control \cite{dna_iros2011,dna_tcst2014}, which takes the control error as a difference of measured force to reference force (10-15N, depending on BMI \cite{Tirindelli2020}).
Meanwhile, for automatic curvature following, the algorithm fuses the network located spinous processes into a continuous path based on the Kalman filter, compensating for the missing parts (intervertebral gaps). It also adjusts the probe's orientation along the normal vector to a subject's back, the speed of orientation change depends on the spinal region, which is predicted by the network described below.

The robot uses a machine-learning algorithm to follow the spinal curvature and detect the spinal region for rotation adjustments, described in earlier stage of the current work \cite{victorova2021follow}. The fully connected network model was designed to detect the spinous process location to indicate the spine center at each ultrasound B-mode frame in real-time. The model presented in Figure \ref{fig:network_two_heads} is build on ResNet\cite{he2015deep} backbone, which is used to extract the feature vector from the input ultrasound image. The model has two heads, first serves as spinous location prediction, where the backbone is followed by three deconvolutional layers with batch normalization \cite{ioffe2015batch} and ReLU activation \cite{krizhevsky2012imagenet}, which serve as a decoder to generate the image features heatmap \cite{xiao2018simple}. The deconvolutional layers are followed by a single 1x1 convolutional layer, which converts the resulting features matrix to a final heatmap where each image pixel intensity represents the probability of being one of the classes; in our case, the maximum intensity represents the high probability of the pixel belonging to a spinous process class. 
The mean squared error (MSE) determines the difference between the actual and forecasted heatmaps. The final location is calculated from the heatmap's greatest intensity pixels.

The second head of the network has a fully connected layer after the backbone. The output is the probability of each ultrasound image belonging to a spinal region class: ``sacrum", ``lumbar," ``thoracic." According to the output class, the algorithm changes the control gain $K_{pitch}$ which specifies how fast the rotation of the probe is changing, $r_{x} = -K_{pitch}m_{x}$, where the torque $m_{x}$ is obtained from the force sensor. For example, the thoracic part of the spine generally has larger curves in the sagittal view than the lumbar part; thus, a larger control gain is needed. 
25,774 images were used for training and testing of the network resulting in a 5-fold cross-validation accuracy of $0.939\pm0.010$ for the classification
task and $0.973\pm0.006$ for the localization task with mean distance error in the range of $0.8-3.3$ mm.


\subsection{Deformity Angle Measurement}

According to the scoliosis application and the nature of the images, the observational criteria for image quality assessment can include a clear spinous process shadow path, which allows unambiguous deformity angle measurement; visibility of the spinal features, such as lamina, transverse process, etc.; uninterrupted acoustic coupling, thus no missing spots due to the contact lost with the skin.
However, these criteria only give an overview of the image quality; the indirect measure of the quality is needed. The quality assessment criteria for image quality for scoliosis assessment is the measurement of the spinal deformity. If the measurements of the robotic approach are the same or comparable with angles obtained by the manual method, the image quality is considered to be good.

In total, 23 subjects were examined with proposed automatic scanning for scoliosis assessment; each had one or two deformity angles. Two raters, R1 and R2, conducted the measurement of spinal curve deformity angle (SPA) according to the method presented in \cite{Zheng2016}. The rater R2 had more than 5 years of experience in measuring scoliosis, while the other one R1 was a newcomer who received basic instructions on the measurement procedure.
 The deformity angle measurement procedure is summarized in Figure \ref{fig:robotic_appr_angle}b. The raters used the software to load coronal projection images and draw the lines at the most tilted vertebrae levels of the spinal curve. Then the software outputs the spinous process deformity angles, calculated as an angle between drawn lines. The raters measured the deformity angle on six coronal images from three robotic and three manual scans.

\subsection{Statistical Analysis}
Statistical analyses were performed using the IBM SPSS Statistics Version 25 (IBM, USA).
To assess the intra-rater reliability of spinal deformity angles measurement, the intra-class correlation coefficient ICC with 95\% confidence interval was calculated. Koo et al., \cite{koo2016guideline} stated the guidelines for proper ICC test design selection. For intra-rater reliability, the usual model used was two-way mixed with absolute agreement definition; the reported value was single-rating since there was no averaging of the measurements. The absolute agreement was used to see how different the same angle measurements were among different scans by the same rater.

Similarly, the inter-rater analysis uses ICC to reflect the correlation between the measurements of two raters; it is performed on means of three scans of each method for each rater; thus, the reported value is average-rating.
According to Portney et al.,\cite{portney2009foundations}, the ICC values between 0.5 and 0.75 indicate moderate reliability, values greater than 0.75 indicate good reliability and greater than 0.9 excellent reliability. Since 5\degree in angles measurement variability is acceptable and commonly reported \cite{langensiepen2013measuring,Zheng2016}, the percentage and number of curves, which MAD is less than 5\degree is reported.


The Wilcoxon signed-rank test was used to compare the angles measured on coronal images obtained from robotic and manual assessments to evaluate the validity of ultrasound assessment.
 The inter-method analysis employed a non-parametric Wilcoxon test instead of the standard t-test method. This test shows whether there is a statistical difference between the angles obtained by robotic and manual scanning. The analysis was performed on the mean of three scans for each scanning method. Another metric to assess the inter-method correlation is $R^2$ value computed with the linear regression method.


\section{Results}

\begin{figure}[]
    \centering
    \includegraphics[width=0.85\linewidth]{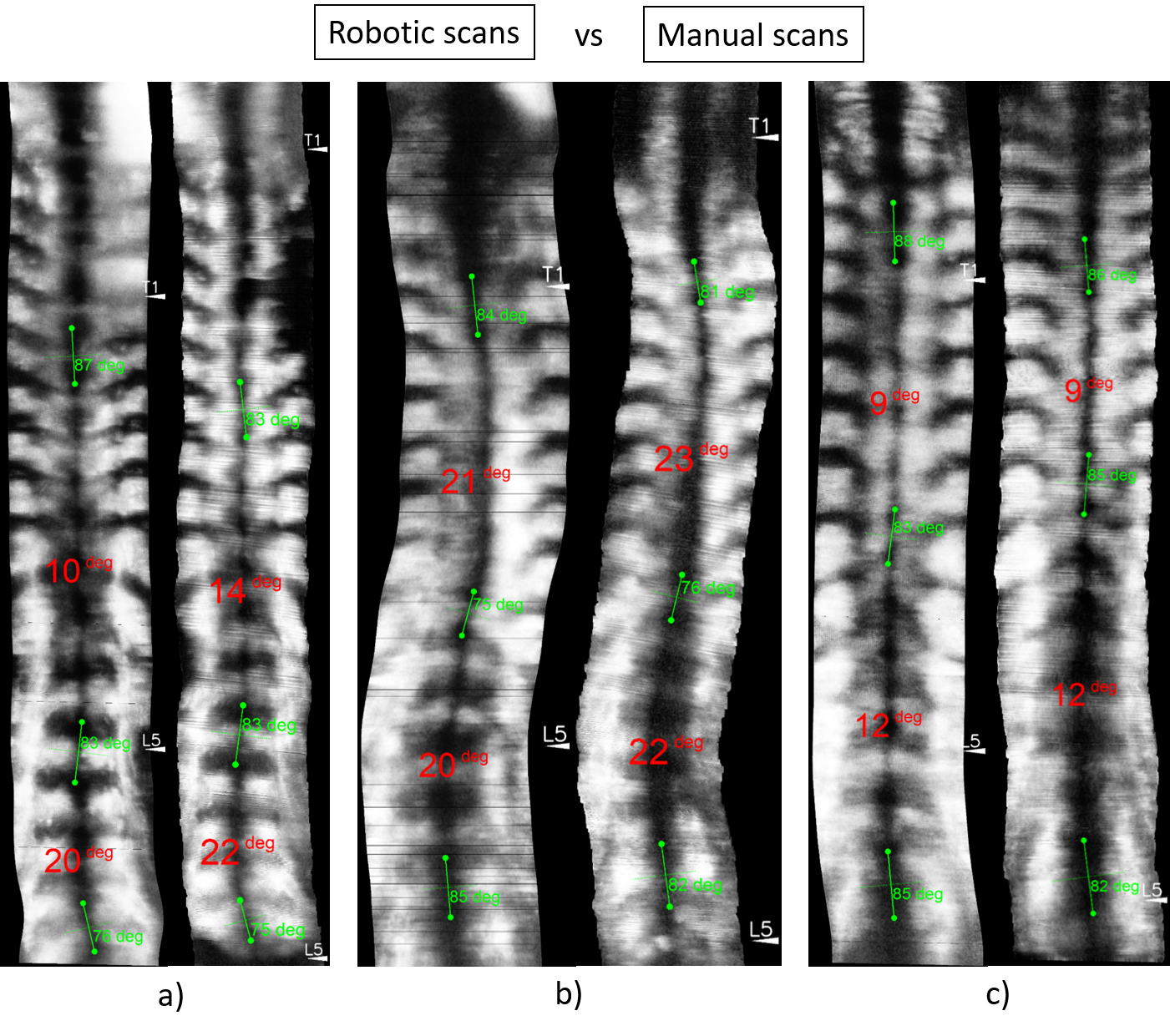}
    \includegraphics[width=0.85\linewidth]{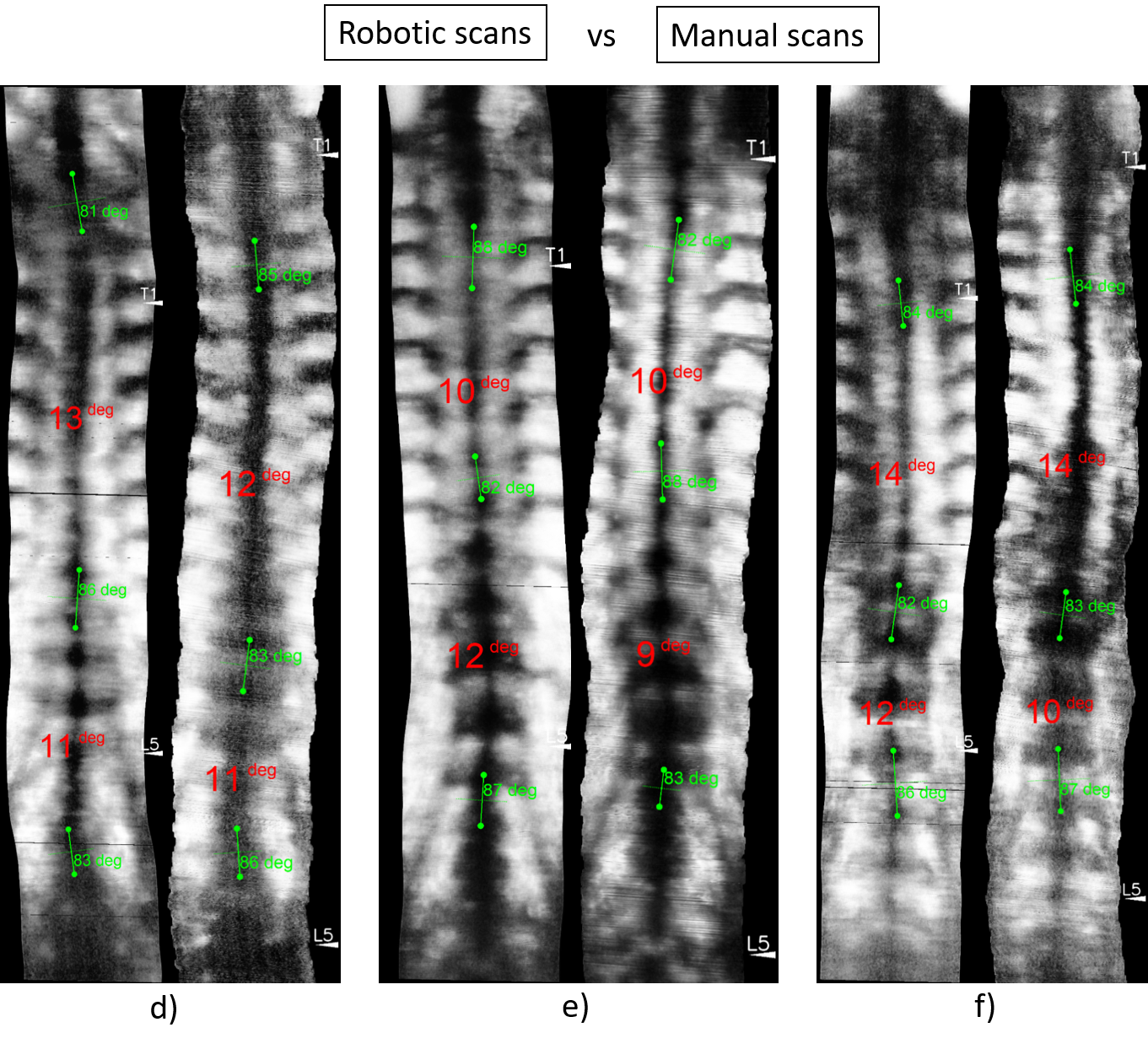}
    \caption[Comparison of robotic and manual scans.]{Comparison of coronal images resulted from robotic (LEFT) and manual (RIGHT) scanning with measured SPA angles for six subjects with different scoliosis severity. The robotic scanning yield smoother coronal reconstructions and more prominent spinal features.}
    \label{fig:robotic vs manual}
\end{figure}

Recruited subjects had an average body mass index (BMI) of $18.9 \pm 2.2$ $kg/m^2$ and age of $15 \pm 1.0$ years. 
Ten subjects had a single thoracic or lumbar curve, while 13 possessed a double curve, yielding a total of 36 curves for analysis in this study. The SPA measured angles of these curves ranged from $7\degree$ to $31\degree$, and the average value was $15.7\degree\pm5.6\degree$. While all the subjects have mild or moderate scoliosis, the second curve of 5 subjects was slightly less than $10\degree$. Since the purpose of the study is the curves angles comparison and not the scoliosis detection, these angles were also included in the study.

Figure \ref{fig:robotic vs manual} presents one robotic (left) and one manual (right) scan for each of six exemplary subjects. It can be noticed that the resulting curve angles are similar between the two scanning approaches.
The survey result shows no significant difference between the compatibility of robotic (7.8 out of 10) and manual (8.7 out of 10) scanning. 
 
Table \ref{tab:intra-rater} displays the ICC values for two-way mixed model with absolute agreement definition, single-rating type. 
The MAD, SD, and SEM values were calculated for each curve based on angles measured on three scans of one method (robotic or manual). The average values (with min and max) for all the curves for intra-rater analysis are presented in Table \ref{tab:intra-rater}. The highest value of ICC=0.868 was found for robotic method Rater 2, where MAD was below $5\degree$ except for one outlier. The lowest value of ICC=0.753 was found for manual method Rater 1 with MAD below $5\degree$ for all angles.


\begin{table}[]
\caption[Intra-rater variation and reliability of coronal curvature assessments between robotic scanning and manual scanning]{\MakeUppercase{Intra-rater variation and reliability of coronal curvature assessments using robotic ultrasound scanning compared with manual ultrasound scanning}}
\label{tab:intra-rater}
\resizebox{\textwidth}{!}{%
\begin{tabular}{lllllll}
\hline
Methods                                                               & Raters                    & $ICC (95\%CI)^a$ &$MAD^b$(max), \degree & \% below 5\degree$^c$  & $SD^d$(max),\degree & $SEM^e$(max), \degree \\ \hline
\rowcolor[HTML]{EFEFEF} 
\multicolumn{1}{c}{\cellcolor[HTML]{EFEFEF}}                          & R1                        & 0.804 (0.690-0.887)             & $2.3\pm1.1 (4.0)$  & 100 (36)                            & $1.8\pm1.0 (4.0)$          & $1.1\pm0.6 (2.3)$           \\
\multicolumn{1}{c}{\multirow{-2}{*}{\cellcolor[HTML]{EFEFEF}Robotic}} & {\color[HTML]{000000} R2} & {\color[HTML]{000000} 0.868 (0.785-0.925)}         &            $2.5\pm1.4 (5.3)$   & 97 (35)          &        $1.9\pm1.1 (4.0)$              &      $1.1\pm0.6 (2.3)$                  \\
\rowcolor[HTML]{EFEFEF} 
\cellcolor[HTML]{EFEFEF}                                              & R1                        & 0.753 (0.619-0.855)              & $2.6\pm1.0 (4.7)$    &100 (36)                           & $2.1\pm0.8 (3.6)$            & $1.2\pm0.5 (2.0)$           \\
\multirow{-2}{*}{\cellcolor[HTML]{EFEFEF}Manual}                      & R2                        &    0.829 (0.725-0.902)                            &   $2.5\pm1.7 (6)$               &     88.8 (32)                     &   $2.0\pm1.3 (4.7)$                     &       $1.1\pm0.7 (2.7)$                \\ \hline
\end{tabular}%

}
\footnotesize{$CI^a$:confidence intervals, $MAD^b$: mean absolute difference;\% below 5\degree$^c$(num of angles): clinically not significant difference of angles below 5\degree;   $SD^d$: standard deviation;  $SEM^e$: standard error of the mean}\\
\end{table}

\begin{figure}[]
    \centering
    \includegraphics[width=\linewidth]{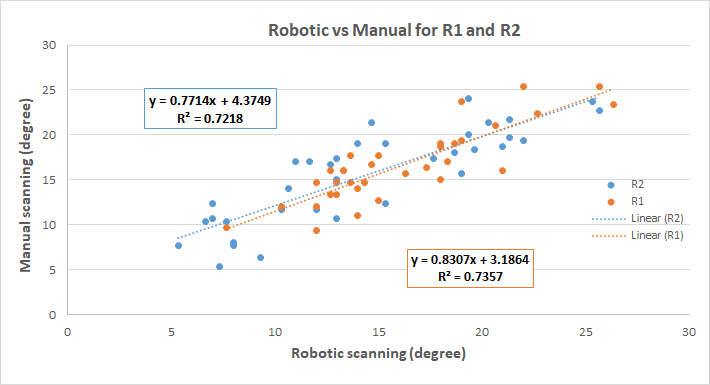}
    \caption[Correlation and equation between the SPA measurement results on robotic scanning and manual scanning]{Correlation and equation between the SPA measurement results by robotic scanning and manual scanning for mean of rater 1 and rater 2.}
    \label{fig:correlation_R^2}
\end{figure}

The results of the inter-method comparison are presented in Table \ref{tab:inter-method}. The Wilcoxon test with the value of 0.166 and 0.06 (with p = 0.05) showed no significant difference between angles measured on images obtained with a robotic system (average of three scans) and with manual scanning. 
The $R^2$ values of 0.73 and 0.72 demonstrated a moderate correlation between the SPA results by the two scanning approaches, robotic and manual. The correlation plot between the two methods is displayed in Figure \ref{fig:correlation_R^2}.
According to Table \ref{tab:inter-rater} inter-rater reliability had ICC=0.772 for robotic and ICC=0.843 for manual methods. The MAD was below $5\degree$ for all cases except four angles for manual scan and eight for robotic.

\begin{table}[]
\caption[Comparison of coronal curvatures assessments between robotic scanning and manual scanning (on mean of three scans for each method).]{\MakeUppercase{Comparison of coronal curvatures assessments between robotic ultrasound scanning and manual ultrasound scanning (on mean of three scans for each method).}}
\label{tab:inter-method}
\resizebox{\textwidth}{!}{%
\begin{tabular}{cccccc}
\hline
Raters & Methods    &$MAD^a$(max), \degree          & \% below 5\degree$^b$                     & Wilcoxon$^c$ (p=0.05) & $R^2$ \\ \hline
\rowcolor[HTML]{EFEFEF} 
R1     & robotic vs manual         & $1.8\pm1.4(5.7)$    &   97(35)        & 0.166             & 0.73  \\
\rowcolor[HTML]{EFEFEF} 
R2     & \cellcolor[HTML]{EFEFEF}robotic vs manual     & $2.5\pm1.6(6.0)$  &  92(33) &    0.06             &    0.72   \\  \hline
\end{tabular}%
}
\footnotesize{$MAD^a$: mean absolute difference;\% below 5\degree$^b$(num of angles): clinically not significant difference of angles below 5\degree;   $SD^d$: standard deviation; Wilcoxon$^c$: a non-parametric signed-rank test}\\

\end{table}

\begin{table}[]
\caption[Inter-rater correlation for each scanning approach]{\MakeUppercase{Inter-rater correlation for robotic ultrasound and manual ultrasound approaches.}}
\label{tab:inter-rater}
\resizebox{\textwidth}{!}{%
\begin{tabular}{ccccccc}
\hline
 Methods &  Raters    & $ICC (95\%CI)^a$ &$MAD^b$(max), \degree & \% below 5\degree$^c$  & $SD^d$(max),\degree & $SEM^e$(max), \degree \\ \hline
\rowcolor[HTML]{EFEFEF} 
Robotic & R1 vs R2    & 0.772 (0.557-0.883) & $3.6\pm2.7(10.6)$ &77.8(28) & $2.5\pm1.9(7.5)$   & $1.8\pm1.3(5.3)$ \\
\rowcolor[HTML]{EFEFEF} 
Manual  & \cellcolor[HTML]{EFEFEF}R1 vs R2 & 0.843 (0.694-0.919) & $2.8\pm2.2(8)$ & 89(32) &$1.9\pm1.6(5.6)$ & $1.4\pm1.1(4.0)$ \\ \hline
\end{tabular}%
}
\footnotesize{$CI^a$:confidence intervals, $MAD^b$: mean absolute difference;\% below 5\degree$^c$(num of angles): clinically not significant difference of angles below 5\degree;   $SD^d$: standard deviation;  $SEM^e$: standard error of the mean}\\
\end{table}

\section{Discussion and Conclusions}


Kottner et al. in \cite{kottner2011guidelines} summarized the interpretation of the ICC values; he showed that the values of 0.6 to 0.8 are often used as a minimum standard for research purposes, where the sample size is relatively smaller, however for clinical practice, the value should be at least 0.9. 

The statistical analysis shows that both robotic and manual approaches have good reliability, with values slightly higher for the robotic approach. According to the Wilcoxon test, both raters agreed that there was no significant difference between robotic and manual angles. A moderate linear correlation was found between robotic and manual approaches; however, the variation in the correlation numbers for the Wilcoxon test can be explained by the considerable difference in the rater's experience.

Previous studies have revealed that Cobb angle measurements of moderate scoliosis on X-ray images have a 3\degree to 5\degree  intra- and inter-observer variability \cite{cobb1948outline, scholten1987analysis,tanure2010reliability}. The intra-class correlation coefficient ICC for Cobb's angle was from 0.83 to 0.99 according to a systematic review \cite{langensiepen2013measuring}. Angle measurements on ultrasound images had a variability of 2\degree to 5\degree \cite{Zheng2016,brink2018reliability,wang2015reliability}.
The current study shows that MAD for inter-rater and intra-rater mostly lays in a range of 0\degree to 5 \degree with a few outliers.

\begin{figure}[]
    \centering
    \includegraphics[width=\linewidth]{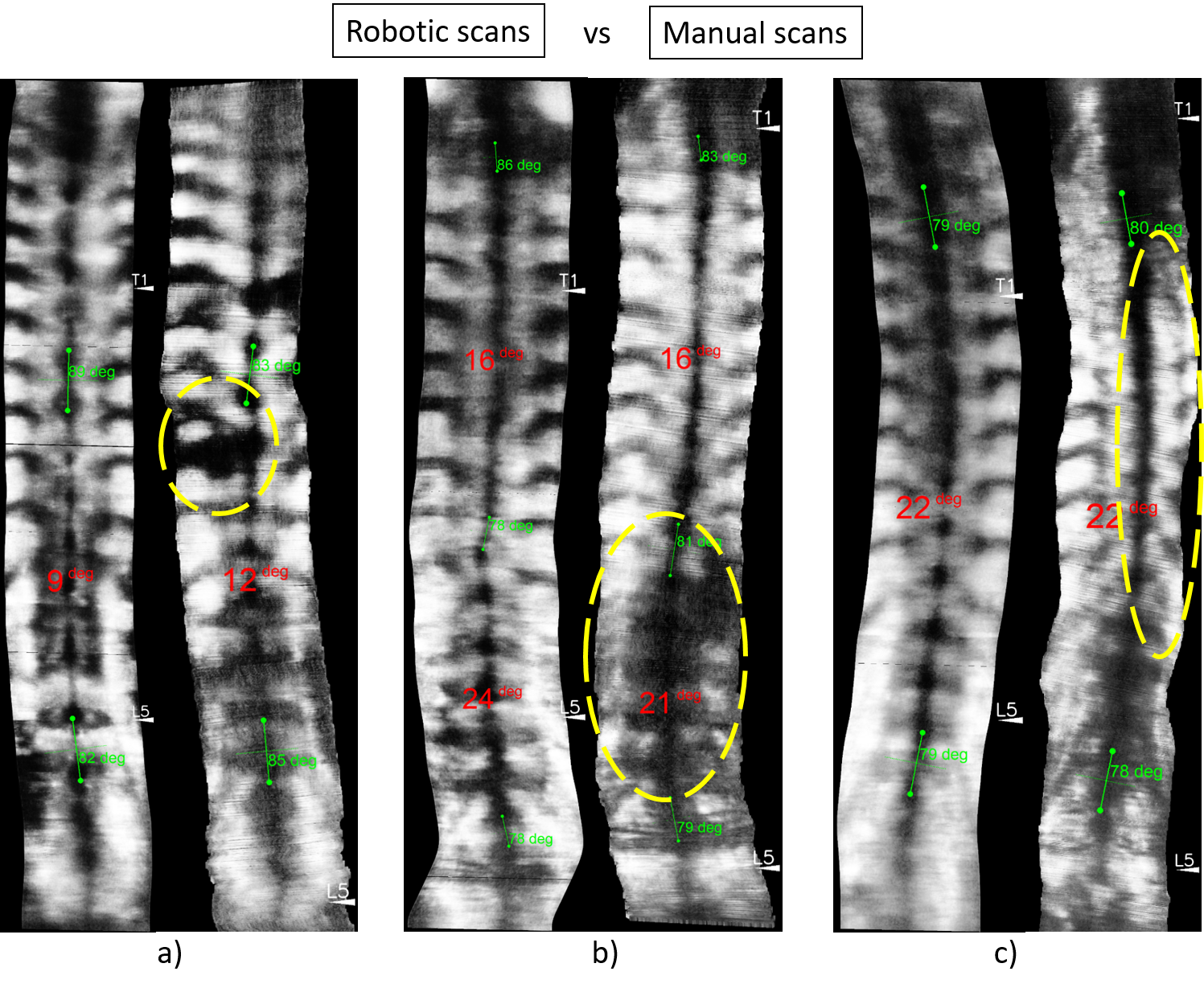}
    \caption[Comparison of robotic and manual coronal images. Challenges of robotic approach.]{Comparison of coronal images obtained from robotic (LEFT) and manual (RIGHT) scanning shows challenges of the manual approach. a) skin contact lost (black spots); b) spinal lumbar features are less prominent for manual scanning; c) the challenges of following the curve leave the spinal aside of field of view.}
    \label{fig: manual_problematic}
\end{figure}

According to a review of medical image quality assessment, \cite{chow2016review}, currently, there is no standardized way to assess ultrasound image quality. Specifically to an application, US image quality can be defined by comparison to other modalities, such as X-ray, CT, MRI. Suppose there are no reference images of different modalities. In that case, the image quality is indirectly defined as an ability to see the features of interest (target anatomical structures) within the image \cite{li2021overview}.

Figure \ref{fig: manual_problematic} illustrates the noticeable benefits of the robotic approach. Figure \ref{fig: manual_problematic}a  shows the more regular contact with the skin surface during robotic scanning compared to manual, where the points of contact lost appeared in black for four scanned subjects. The other common observation is that the features of the lumbar spine were less prominent and poorly distinguishable for ten subjects scanned manually compared to robotic scanning, as shown in Figure \ref{fig: manual_problematic}b. These advantages of the robotic approach can be explained by the stable force control, which includes the constant pressure onto the skin and probe's orientation adjustments.   
The other example (Figure \ref{fig: manual_problematic}c) shows the benefits of having a spinal curvature tracking algorithm to center the spine in the field of view, which ensures that the spine is always in the field of view (which is especially useful for patients with severe scoliosis) and also gives a better view on the surrounding features such as laminae and transverse processes to utilize different approaches for angle measurements. For five subjects scanned manually, the spinous process surrounding features were out of the field of view compared to robotic scanning.
In general, the generated coronal image of the robotic scan looks smoother, the spine is lying in the middle of the image, and bony features are more distinguishable than in the manual scan.

The number of US frames for reconstruction is similar in both robotic and manual approaches; the scanning speed in the robotic approach is twice slower than for manual. However, the speed difference may give an additional smoothness contour to the reconstructed image in the robotic scan; the concern is that a longer scanning time may cause instability in the subject's posture since the subject might make small motions during that time.

\begin{figure}[]
    \centering
    \includegraphics[width=\linewidth]{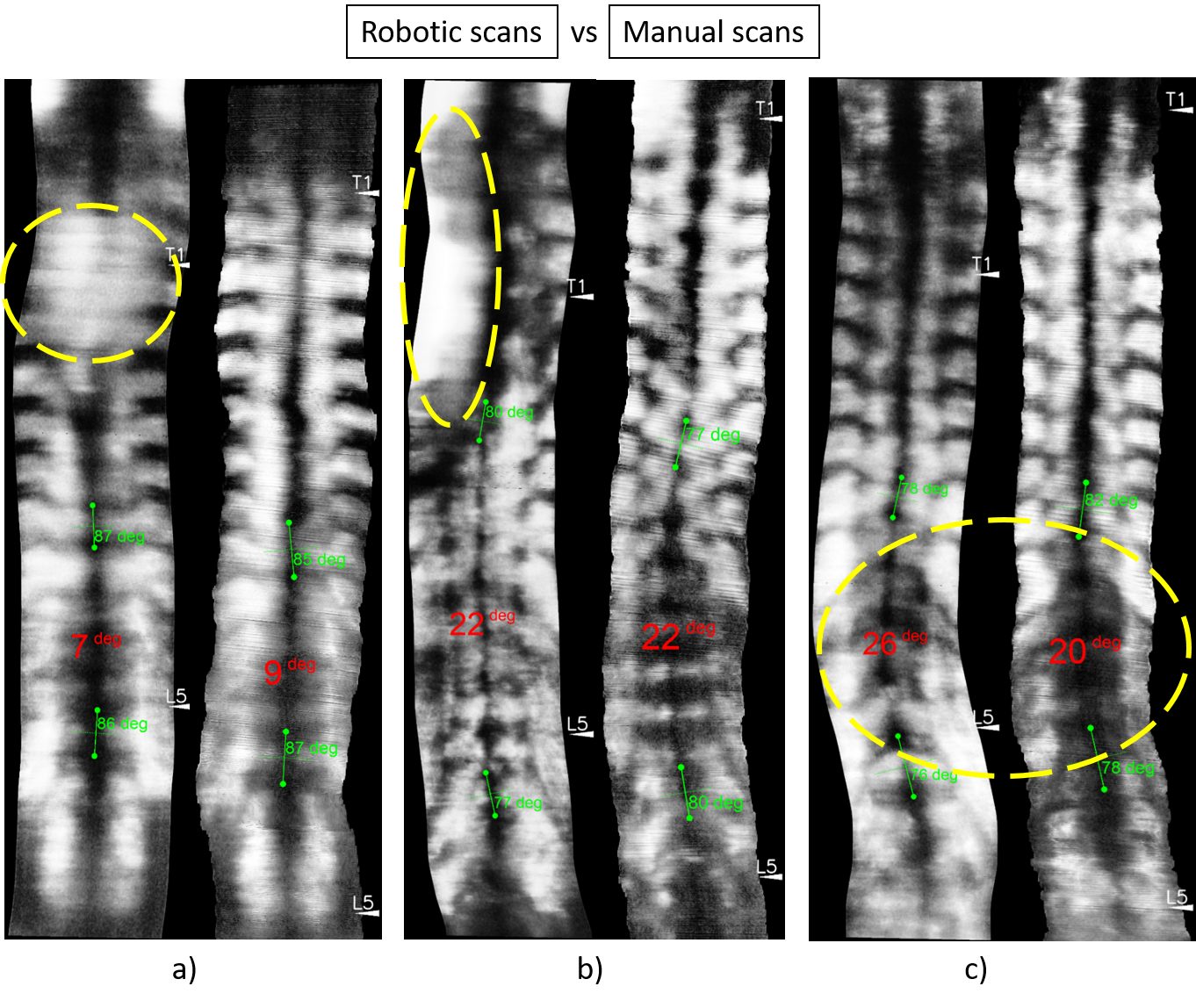}
    \caption[Comparison of robotic and manual coronal images. Challenges of robotic approach.]{Comparison of coronal images obtained from robotic (LEFT) and manual (RIGHT) scanning show challenges of the robotic approach. a) a subject with narrow space between scapula bones b) a subject with uneven back in coronal plane c) a subject bent his/her back under the scanning pressure due to the weak muscular system (a tall subject with low BMI). }
    \label{fig:robotic vs manual_problematic}
\end{figure}

Several limitations for using the robotic approach were encountered during the study. Figure \ref{fig:robotic vs manual_problematic} illustrates less successful robotic scanning results. Coronal image of robotic scan for the first case (Figure \ref{fig:robotic vs manual_problematic}a) represents the subject with low BMI and tight prominent scapula bones, which leaves too narrow space between the right and left scapula for a robotic actuated probe to pass through. Thus, the probe detached from the skin at the scapula level during the scanning, and image quality suffered in that region. In comparison, the manual scanning handled the tight scapula situation well since the operator could roll the probe to fit the narrow space and pass through. The issue with tight scapula was encountered in two subjects out of a total number of examined subjects. The robotic system cannot recognize the tight scapula cases and perform the necessary roll rotation. 

The second challenging case for robotic scanning is displayed in Figure \ref{fig:robotic vs manual_problematic}b. The subject had an uneven back, with high concavity at the left side of the back at the level of the thoracic region; thus, the left side of the probe was not compliant with the subject's back. There was only one subject with this peculiarity during the experimental trials. For such cases, two solutions can be used. The first one is to apply a greater amount of gel to fill in the gap; the second one is the greater robotic row rotation, which was implemented by a force sensing approach; however, this type of rotation was not fully studied since there were no subjects with severe scoliosis.  

The third case shows a tall subject with low BMI (Figure \ref{fig:robotic vs manual_problematic}c), where the subject could not resist the force applied by the robot (even the lowest settings of 10 N) and bend in the lumbar region as the probe was applying pressure on the back. Thus, the lumbar curve from a robotic scan is much larger than that from a human scan. This situation was observed for three subjects out of total. It was noticed that tall subjects with low BMI tended to have this problem. This might be because they had a weaker muscular system.

However, the study already shows a good correlation among the angles measured from robotic and manual scanning; it is suggested to do a more extensive case study on more subjects. Manual scans were performed by an expert who has done this procedure for more than five years. To train an operator, in general, it takes about six months to learn the scanning procedure on different types of subjects, to perform multiple scans for each to learn to correct the scanning path. In Hong Kong, the six-month labor cost can be about 15,000 USD for one operator, while the robotic arm price starts from 10,000 USD + software. It would also be meaningful to compare the robotic approach with a non-experienced user, which might show that the robotic approach performs better. 
For comparison purposes, it is better to obtain the ground truth spinal image from the patient before scanning, such as X-ray or Scolioscan image, which proved to be reliable by previous study \cite{Zheng2016}.

The comparison of robotic and manual ultrasound scanning for scoliosis evaluation was given in this study. The results of this study show that scoliosis deformity angles measured on images produced using a robotic approach are comparable to those obtained with manual ultrasound scanning. The benefits of employing a robotic approach were explored. This study paves the way for further research into the robotic approach on a larger group of participants with varying degrees of scoliosis severity, as well as eventual commercialization.

\subsubsection*{Acknowledgments} The authors thank Stella Chui Yi Chan and Kelly Lai for their contribution in managing the recruitment of the subjects. 

\subsubsection*{Conflicts of Interests Statement}
The authors have no conflicts of interest to declare.


\bibliography{sn-bibliography.bib}


\begin{thebibliography}{31}
\ifx \bisbn   \undefined \def \bisbn  #1{ISBN #1}\fi
\ifx \binits  \undefined \def \binits#1{#1}\fi
\ifx \bauthor  \undefined \def \bauthor#1{#1}\fi
\ifx \batitle  \undefined \def \batitle#1{#1}\fi
\ifx \bjtitle  \undefined \def \bjtitle#1{#1}\fi
\ifx \bvolume  \undefined \def \bvolume#1{\textbf{#1}}\fi
\ifx \byear  \undefined \def \byear#1{#1}\fi
\ifx \bissue  \undefined \def \bissue#1{#1}\fi
\ifx \bfpage  \undefined \def \bfpage#1{#1}\fi
\ifx \blpage  \undefined \def \blpage #1{#1}\fi
\ifx \burl  \undefined \def \burl#1{\textsf{#1}}\fi
\ifx \doiurl  \undefined \def \doiurl#1{\url{https://doi.org/#1}}\fi
\ifx \betal  \undefined \def \betal{\textit{et al.}}\fi
\ifx \binstitute  \undefined \def \binstitute#1{#1}\fi
\ifx \binstitutionaled  \undefined \def \binstitutionaled#1{#1}\fi
\ifx \bctitle  \undefined \def \bctitle#1{#1}\fi
\ifx \beditor  \undefined \def \beditor#1{#1}\fi
\ifx \bpublisher  \undefined \def \bpublisher#1{#1}\fi
\ifx \bbtitle  \undefined \def \bbtitle#1{#1}\fi
\ifx \bedition  \undefined \def \bedition#1{#1}\fi
\ifx \bseriesno  \undefined \def \bseriesno#1{#1}\fi
\ifx \blocation  \undefined \def \blocation#1{#1}\fi
\ifx \bsertitle  \undefined \def \bsertitle#1{#1}\fi
\ifx \bsnm \undefined \def \bsnm#1{#1}\fi
\ifx \bsuffix \undefined \def \bsuffix#1{#1}\fi
\ifx \bparticle \undefined \def \bparticle#1{#1}\fi
\ifx \barticle \undefined \def \barticle#1{#1}\fi
\bibcommenthead
\ifx \bconfdate \undefined \def \bconfdate #1{#1}\fi
\ifx \botherref \undefined \def \botherref #1{#1}\fi
\ifx \url \undefined \def \url#1{\textsf{#1}}\fi
\ifx \bchapter \undefined \def \bchapter#1{#1}\fi
\ifx \bbook \undefined \def \bbook#1{#1}\fi
\ifx \bcomment \undefined \def \bcomment#1{#1}\fi
\ifx \oauthor \undefined \def \oauthor#1{#1}\fi
\ifx \citeauthoryear \undefined \def \citeauthoryear#1{#1}\fi
\ifx \endbibitem  \undefined \def \endbibitem {}\fi
\ifx \bconflocation  \undefined \def \bconflocation#1{#1}\fi
\ifx \arxivurl  \undefined \def \arxivurl#1{\textsf{#1}}\fi
\csname PreBibitemsHook\endcsname

\bibitem{cobb1948outline}
\begin{barticle}
\bauthor{\bsnm{Cobb}, \binits{J.}}:
\batitle{Outline for the study of scoliosis}.
\bjtitle{Instr Course Lect AAOS}
\bvolume{5},
\bfpage{261}--\blpage{275}
(\byear{1948})
\end{barticle}
\endbibitem

\bibitem{hoffman1989breast}
\begin{barticle}
\bauthor{\bsnm{Hoffman}, \binits{D.A.}},
\bauthor{\bsnm{Lonstein}, \binits{J.E.}},
\bauthor{\bsnm{Morin}, \binits{M.M.}},
\bauthor{\bsnm{Visscher}, \binits{W.}},
\bauthor{\bsnm{Harris~III}, \binits{B.S.}},
\bauthor{\bsnm{Boice~Jr}, \binits{J.D.}}:
\batitle{Breast cancer in women with scoliosis exposed to multiple diagnostic x
  rays}.
\bjtitle{JNCI: Journal of the National Cancer Institute}
\bvolume{81}(\bissue{17}),
\bfpage{1307}--\blpage{1312}
(\byear{1989})
\end{barticle}
\endbibitem

\bibitem{levy1994projecting}
\begin{barticle}
\bauthor{\bsnm{Levy}, \binits{A.R.}},
\bauthor{\bsnm{Goldberg}, \binits{M.S.}},
\bauthor{\bsnm{Hanley}, \binits{J.A.}},
\bauthor{\bsnm{Mayo}, \binits{N.E.}},
\bauthor{\bsnm{Poitras}, \binits{B.}}:
\batitle{Projecting the lifetime risk of cancer from exposure to diagnostic
  ionizing radiation for adolescent idiopathic scoliosis.}
\bjtitle{Health physics}
\bvolume{66}(\bissue{6}),
\bfpage{621}--\blpage{633}
(\byear{1994})
\end{barticle}
\endbibitem

\bibitem{Zheng2016}
\begin{botherref}
\oauthor{\bsnm{Zheng}, \binits{Y.-p.}},
\oauthor{\bsnm{Lee}, \binits{T.T.-y.}},
\oauthor{\bsnm{Lai}, \binits{K.K.-l.}},
\oauthor{\bsnm{Yip}, \binits{B.H.-k.}},
\oauthor{\bsnm{Zhou}, \binits{G.-q.}},
\oauthor{\bsnm{Jiang}, \binits{W.-w.}},
\oauthor{\bsnm{Cheung}, \binits{J.C.-w.}},
\oauthor{\bsnm{Wong}, \binits{M.-s.}},
\oauthor{\bsnm{Ng}, \binits{B.K.-w.}},
\oauthor{\bsnm{Cheng}, \binits{J.C.-y.}}:
{A reliability and validity study for Scolioscan: a radiation-free scoliosis
  assessment system using 3D ultrasound imaging}.
Scoliosis and Spinal Disorders,
1--15
(2016).
\doiurl{10.1186/s13013-016-0074-y}
\end{botherref}
\endbibitem

\bibitem{wang2015reliability}
\begin{barticle}
\bauthor{\bsnm{Wang}, \binits{Q.}},
\bauthor{\bsnm{Li}, \binits{M.}},
\bauthor{\bsnm{Lou}, \binits{E.H.}},
\bauthor{\bsnm{Wong}, \binits{M.S.}}:
\batitle{Reliability and validity study of clinical ultrasound imaging on
  lateral curvature of adolescent idiopathic scoliosis}.
\bjtitle{PloS one}
\bvolume{10}(\bissue{8}),
\bfpage{0135264}
(\byear{2015})
\end{barticle}
\endbibitem

\bibitem{brink2018reliability}
\begin{barticle}
\bauthor{\bsnm{Brink}, \binits{R.C.}},
\bauthor{\bsnm{Wijdicks}, \binits{S.P.}},
\bauthor{\bsnm{Tromp}, \binits{I.N.}},
\bauthor{\bsnm{Schl{\"o}sser}, \binits{T.P.}},
\bauthor{\bsnm{Kruyt}, \binits{M.C.}},
\bauthor{\bsnm{Beek}, \binits{F.J.}},
\bauthor{\bsnm{Castelein}, \binits{R.M.}}:
\batitle{A reliability and validity study for different coronal angles using
  ultrasound imaging in adolescent idiopathic scoliosis}.
\bjtitle{The Spine Journal}
\bvolume{18}(\bissue{6}),
\bfpage{979}--\blpage{985}
(\byear{2018})
\end{barticle}
\endbibitem

\bibitem{chen2012ultrasound}
\begin{botherref}
\oauthor{\bsnm{Chen}, \binits{W.}},
\oauthor{\bsnm{Le}, \binits{L.H.}},
\oauthor{\bsnm{Lou}, \binits{E.H.}}:
Ultrasound imaging of spinal vertebrae to study scoliosis
(2012)
\end{botherref}
\endbibitem

\bibitem{scolioAirlai2021}
\begin{barticle}
\bauthor{\bsnm{Lai}, \binits{K.K.-L.}},
\bauthor{\bsnm{Lee}, \binits{T.T.-Y.}},
\bauthor{\bsnm{Lee}, \binits{M.K.-S.}},
\bauthor{\bsnm{Hui}, \binits{J.C.-H.}},
\bauthor{\bsnm{Zheng}, \binits{Y.-P.}}:
\batitle{Validation of scolioscan air-portable radiation-free three-dimensional
  ultrasound imaging assessment system for scoliosis}.
\bjtitle{Sensors}
\bvolume{21}(\bissue{8}),
\bfpage{2858}
(\byear{2021})
\end{barticle}
\endbibitem

\bibitem{Cheung2015}
\begin{botherref}
\oauthor{\bsnm{Cheung}, \binits{C.-w.J.}},
\oauthor{\bsnm{Zhou}, \binits{G.-q.}},
\oauthor{\bsnm{Law}, \binits{S.-y.}},
\oauthor{\bsnm{Mak}, \binits{T.-m.}},
\oauthor{\bsnm{Lai}, \binits{K.-l.}}:
{Ultrasound Volume Projection Imaging for Assessment of Scoliosis}
\textbf{0062}(c)
(2015).
\doiurl{10.1109/TMI.2015.2390233}
\end{botherref}
\endbibitem

\bibitem{Jiang2019}
\begin{barticle}
\bauthor{\bparticle{wei} \bsnm{Jiang}, \binits{W.}},
\bauthor{\bparticle{quan} \bsnm{Zhou}, \binits{G.}},
\bauthor{\bsnm{Lai}, \binits{K.L.}},
\bauthor{\bparticle{yu} \bsnm{Hu}, \binits{S.}},
\bauthor{\bparticle{yu} \bsnm{Gao}, \binits{Q.}},
\bauthor{\bparticle{yan} \bsnm{Wang}, \binits{X.}},
\bauthor{\bparticle{ping} \bsnm{Zheng}, \binits{Y.}}:
\batitle{{A fast 3-D ultrasound projection imaging method for scoliosis
  assessment}}.
\bjtitle{Mathematical Biosciences and Engineering}
\bvolume{16}(\bissue{3}),
\bfpage{1067}--\blpage{1081}
(\byear{2019}).
\doiurl{10.3934/mbe.2019051}
\end{barticle}
\endbibitem

\bibitem{vo2019semi}
\begin{barticle}
\bauthor{\bsnm{Vo}, \binits{Q.N.}},
\bauthor{\bsnm{Le}, \binits{L.H.}},
\bauthor{\bsnm{Lou}, \binits{E.}}:
\batitle{A semi-automatic 3d ultrasound reconstruction method to assess the
  true severity of adolescent idiopathic scoliosis}.
\bjtitle{Medical \& biological engineering \& computing}
\bvolume{57}(\bissue{10}),
\bfpage{2115}--\blpage{2128}
(\byear{2019})
\end{barticle}
\endbibitem

\bibitem{Victorova2019}
\begin{bchapter}
\bauthor{\bsnm{{Victorova}}, \binits{M.}},
\bauthor{\bsnm{{Navarro-Alarcon}}, \binits{D.}},
\bauthor{\bsnm{{Zheng}}, \binits{Y.}}:
\bctitle{3d ultrasound imaging of scoliosis with force-sensitive robotic
  scanning}.
In: \bbtitle{2019 Third IEEE International Conference on Robotic Computing
  (IRC)},
pp. \bfpage{262}--\blpage{265}
(\byear{2019}).
\doiurl{10.1109/IRC.2019.00049}
\end{bchapter}
\endbibitem

\bibitem{Tirindelli2020}
\begin{barticle}
\bauthor{\bsnm{{Tirindelli}}, \binits{M.}},
\bauthor{\bsnm{{Victorova}}, \binits{M.}},
\bauthor{\bsnm{{Esteban}}, \binits{J.}},
\bauthor{\bsnm{{Kim}}, \binits{S.T.}},
\bauthor{\bsnm{{Navarro-Alarcon}}, \binits{D.}},
\bauthor{\bsnm{{Zheng}}, \binits{Y.P.}},
\bauthor{\bsnm{{Navab}}, \binits{N.}}:
\batitle{Force-ultrasound fusion: Bringing spine robotic-us to the next
  “level”}.
\bjtitle{IEEE Robotics and Automation Letters}
\bvolume{5}(\bissue{4}),
\bfpage{5661}--\blpage{5668}
(\byear{2020})
\end{barticle}
\endbibitem

\bibitem{Huang2021}
\begin{barticle}
\bauthor{\bsnm{Yang}, \binits{C.}},
\bauthor{\bsnm{Jiang}, \binits{M.}},
\bauthor{\bsnm{Chen}, \binits{M.}},
\bauthor{\bsnm{Fu}, \binits{M.}},
\bauthor{\bsnm{Li}, \binits{J.}},
\bauthor{\bsnm{Huang}, \binits{Q.}}:
\batitle{Automatic 3-d imaging and measurement of human spines with a robotic
  ultrasound system}.
\bjtitle{IEEE Transactions on Instrumentation and Measurement}
\bvolume{70},
\bfpage{1}--\blpage{13}
(\byear{2021}).
\doiurl{10.1109/TIM.2021.3085110}
\end{barticle}
\endbibitem

\bibitem{li2021image}
\begin{botherref}
\oauthor{\bsnm{Li}, \binits{K.}},
\oauthor{\bsnm{Xu}, \binits{Y.}},
\oauthor{\bsnm{Wang}, \binits{J.}},
\oauthor{\bsnm{Ni}, \binits{D.}},
\oauthor{\bsnm{Liu}, \binits{L.}},
\oauthor{\bsnm{Meng}, \binits{M.Q.-H.}}:
Image-guided navigation of a robotic ultrasound probe for autonomous spinal
  sonography using a shadow-aware dual-agent framework.
arXiv preprint arXiv:2111.02167
(2021)
\end{botherref}
\endbibitem

\bibitem{victorova2021follow}
\begin{botherref}
\oauthor{\bsnm{Victorova}, \binits{M.}},
\oauthor{\bsnm{Lee}, \binits{M.K.-S.}},
\oauthor{\bsnm{Navarro-Alarcon}, \binits{D.}},
\oauthor{\bsnm{Zheng}, \binits{Y.}}:
Follow the curve: Robotic-ultrasound navigation with learning based
  localization of spinous processes for scoliosis assessment.
arXiv preprint arXiv:2109.05196
(2021)
\end{botherref}
\endbibitem

\bibitem{Karlsson2005_slam}
\begin{bchapter}
\bauthor{\bsnm{Karlsson}, \binits{N.}},
\bauthor{\bsnm{Bernardo}, \binits{E.}},
\bauthor{\bsnm{Ostrowski}, \binits{J.}},
\bauthor{\bsnm{Goncalves}, \binits{L.}},
\bauthor{\bsnm{Pirjanian}, \binits{P.}},
\bauthor{\bsnm{Munich}, \binits{M.}}:
\bctitle{The vslam algorithm for robust localization and mapping.},
pp. \bfpage{24}--\blpage{29}
(\byear{2005}).
\doiurl{10.1109/ROBOT.2005.1570091}
\end{bchapter}
\endbibitem

\bibitem{dna_iros2011}
\begin{bchapter}
\bauthor{\bsnm{Navarro-Alarcon}, \binits{D.}},
\bauthor{\bsnm{Li}, \binits{P.}},
\bauthor{\bsnm{Yip}, \binits{H.M.}}:
\bctitle{Energy shaping control for robot manipulators in explicit force
  regulation tasks with elastic environments}.
In: \bbtitle{2011 IEEE/RSJ International Conference on Intelligent Robots and
  Systems},
pp. \bfpage{4222}--\blpage{4228}
(\byear{2011}).
\doiurl{10.1109/IROS.2011.6094641}
\end{bchapter}
\endbibitem

\bibitem{dna_tcst2014}
\begin{barticle}
\bauthor{\bsnm{Navarro-Alarcon}, \binits{D.}},
\bauthor{\bsnm{Liu}, \binits{Y.-H.}},
\bauthor{\bsnm{Romero}, \binits{J.G.}},
\bauthor{\bsnm{Li}, \binits{P.}}:
\batitle{Energy shaping methods for asymptotic force regulation of compliant
  mechanical systems}.
\bjtitle{IEEE Transactions on Control Systems Technology}
\bvolume{22}(\bissue{6}),
\bfpage{2376}--\blpage{2383}
(\byear{2014}).
\doiurl{10.1109/TCST.2014.2309659}
\end{barticle}
\endbibitem

\bibitem{he2015deep}
\begin{bchapter}
\bauthor{\bsnm{He}, \binits{K.}},
\bauthor{\bsnm{Zhang}, \binits{X.}},
\bauthor{\bsnm{Ren}, \binits{S.}},
\bauthor{\bsnm{Sun}, \binits{J.}}:
\bctitle{Deep residual learning for image recognition}.
In: \bbtitle{Proceedings of the IEEE Conference on Computer Vision and Pattern
  Recognition},
pp. \bfpage{770}--\blpage{778}
(\byear{2016})
\end{bchapter}
\endbibitem

\bibitem{ioffe2015batch}
\begin{bchapter}
\bauthor{\bsnm{Ioffe}, \binits{S.}},
\bauthor{\bsnm{Szegedy}, \binits{C.}}:
\bctitle{Batch normalization: Accelerating deep network training by reducing
  internal covariate shift}.
In: \bbtitle{International Conference on Machine Learning},
pp. \bfpage{448}--\blpage{456}
(\byear{2015}).
\bcomment{PMLR}
\end{bchapter}
\endbibitem

\bibitem{krizhevsky2012imagenet}
\begin{barticle}
\bauthor{\bsnm{Krizhevsky}, \binits{A.}},
\bauthor{\bsnm{Sutskever}, \binits{I.}},
\bauthor{\bsnm{Hinton}, \binits{G.E.}}:
\batitle{Imagenet classification with deep convolutional neural networks}.
\bjtitle{Advances in neural information processing systems}
\bvolume{25},
\bfpage{1097}--\blpage{1105}
(\byear{2012})
\end{barticle}
\endbibitem

\bibitem{xiao2018simple}
\begin{bchapter}
\bauthor{\bsnm{Xiao}, \binits{B.}},
\bauthor{\bsnm{Wu}, \binits{H.}},
\bauthor{\bsnm{Wei}, \binits{Y.}}:
\bctitle{Simple baselines for human pose estimation and tracking}.
In: \bbtitle{Proceedings of the European Conference on Computer Vision (ECCV)},
pp. \bfpage{466}--\blpage{481}
(\byear{2018})
\end{bchapter}
\endbibitem

\bibitem{koo2016guideline}
\begin{barticle}
\bauthor{\bsnm{Koo}, \binits{T.K.}},
\bauthor{\bsnm{Li}, \binits{M.Y.}}:
\batitle{A guideline of selecting and reporting intraclass correlation
  coefficients for reliability research}.
\bjtitle{Journal of chiropractic medicine}
\bvolume{15}(\bissue{2}),
\bfpage{155}--\blpage{163}
(\byear{2016})
\end{barticle}
\endbibitem

\bibitem{portney2009foundations}
\begin{botherref}
\oauthor{\bsnm{Portney}, \binits{L.G.}},
\oauthor{\bsnm{Watkins}, \binits{M.P.}}, et al.:
Foundations of clinical research: applications to practice
\textbf{892},
11--15
(2009)
\end{botherref}
\endbibitem

\bibitem{langensiepen2013measuring}
\begin{barticle}
\bauthor{\bsnm{Langensiepen}, \binits{S.}},
\bauthor{\bsnm{Semler}, \binits{O.}},
\bauthor{\bsnm{Sobottke}, \binits{R.}},
\bauthor{\bsnm{Fricke}, \binits{O.}},
\bauthor{\bsnm{Franklin}, \binits{J.}},
\bauthor{\bsnm{Sch{\"o}nau}, \binits{E.}},
\bauthor{\bsnm{Eysel}, \binits{P.}}:
\batitle{Measuring procedures to determine the cobb angle in idiopathic
  scoliosis: a systematic review}.
\bjtitle{European Spine Journal}
\bvolume{22}(\bissue{11}),
\bfpage{2360}--\blpage{2371}
(\byear{2013})
\end{barticle}
\endbibitem

\bibitem{kottner2011guidelines}
\begin{barticle}
\bauthor{\bsnm{Kottner}, \binits{J.}},
\bauthor{\bsnm{Audig{\'e}}, \binits{L.}},
\bauthor{\bsnm{Brorson}, \binits{S.}},
\bauthor{\bsnm{Donner}, \binits{A.}},
\bauthor{\bsnm{Gajewski}, \binits{B.J.}},
\bauthor{\bsnm{Hr{\'o}bjartsson}, \binits{A.}},
\bauthor{\bsnm{Roberts}, \binits{C.}},
\bauthor{\bsnm{Shoukri}, \binits{M.}},
\bauthor{\bsnm{Streiner}, \binits{D.L.}}:
\batitle{Guidelines for reporting reliability and agreement studies (grras)
  were proposed}.
\bjtitle{International journal of nursing studies}
\bvolume{48}(\bissue{6}),
\bfpage{661}--\blpage{671}
(\byear{2011})
\end{barticle}
\endbibitem

\bibitem{scholten1987analysis}
\begin{barticle}
\bauthor{\bsnm{Scholten}, \binits{P.}},
\bauthor{\bsnm{Veldhuizen}, \binits{A.}}:
\batitle{Analysis of cobb angle measurements in scoliosis}.
\bjtitle{Clinical Biomechanics}
\bvolume{2}(\bissue{1}),
\bfpage{7}--\blpage{13}
(\byear{1987})
\end{barticle}
\endbibitem

\bibitem{tanure2010reliability}
\begin{barticle}
\bauthor{\bsnm{Tanure}, \binits{M.C.}},
\bauthor{\bsnm{Pinheiro}, \binits{A.P.}},
\bauthor{\bsnm{Oliveira}, \binits{A.S.}}:
\batitle{Reliability assessment of cobb angle measurements using manual and
  digital methods}.
\bjtitle{The Spine Journal}
\bvolume{10}(\bissue{9}),
\bfpage{769}--\blpage{774}
(\byear{2010})
\end{barticle}
\endbibitem

\bibitem{chow2016review}
\begin{barticle}
\bauthor{\bsnm{Chow}, \binits{L.S.}},
\bauthor{\bsnm{Paramesran}, \binits{R.}}:
\batitle{Review of medical image quality assessment}.
\bjtitle{Biomedical signal processing and control}
\bvolume{27},
\bfpage{145}--\blpage{154}
(\byear{2016})
\end{barticle}
\endbibitem

\bibitem{li2021overview}
\begin{botherref}
\oauthor{\bsnm{Li}, \binits{K.}},
\oauthor{\bsnm{Xu}, \binits{Y.}},
\oauthor{\bsnm{Meng}, \binits{M.Q.-H.}}:
An overview of systems and techniques for autonomous robotic ultrasound
  acquisitions.
IEEE Transactions on Medical Robotics and Bionics
(2021)
\end{botherref}
\endbibitem

\end{thebibliography}


\end{document}